\documentclass{article}%
\usepackage{amsmath}
\usepackage{amsfonts}
\usepackage{amssymb}
\usepackage{mathtools}
\usepackage{graphicx}
\usepackage{algorithm}
\usepackage{algpseudocode}%
\providecommand{\U}[1]{\protect\rule{.1in}{.1in}}

\begin{document}

\title{Multiple Instance Learning with Trainable Decision Tree Ensembles}
\author{Andrei V. Konstantinov and Lev V. Utkin,\\Peter the Great St.Petersburg Polytechnic University\\St.Petersburg, Russia\\e-mail: andrue.konst@gmail.com, lev.utkin@gmail.com}
\date{}
\maketitle

\begin{abstract}
A new random forest based model for solving the Multiple Instance Learning
(MIL) problem under small tabular data, called Soft Tree Ensemble MIL
(STE-MIL), is proposed. A new type of soft decision trees is considered, which
is similar to the well-known soft oblique trees, but with a smaller number of
trainable parameters. In order to train the trees, it is proposed to convert
them into neural networks of a specific form, which approximate the tree
functions. It is also proposed to aggregate the instance and bag embeddings
(output vectors) by using the attention mechanism. The whole STE-MIL model,
including soft decision trees, neural networks, the attention mechanism and a
classifier, is trained in an end-to-end manner. Numerical experiments with
tabular datasets illustrate STE-MIL. The corresponding code implementing the
model is publicly available.

\textit{Keywords}: multiple instance learning, decision tree, oblique tree,
random forest, attention mechanism, neural network

\end{abstract}

\section{Introduction}

Many machine learning real-life applications deal with labeled objects called
bags, which consist of several instances such that individual labels of the
instances contained in the bags are not provided, for example, in
histopathology, the histology images can be viewed as bags and its patches
(cells) as instances of the bag
\cite{Hagele-etal-20,Laak-etal-21,Yamamoto-etal-2019}. One can find many
similar application examples, for example, the drug activity prediction
\cite{Dietterich-etal-97}, detecting the lung cancer \cite{Zhu-etal-08}, the
protein function annotation \cite{Wei-etal-19}. A useful frameworks for
modeling the applications is the Multiple Instance Learning (MIL) which can be
regarded as a type of weakly supervised learning
\cite{Amores-13,Babenko-08,Carbonneau-etal-18,Cheplygina-etal-19,Quellec-etal-17,Yao-Zhu-etal-20,Zhou-04}%
. A goal of MIL is to classify new bags based on training data consisting of a
set of labeled bag and to assign labels to unlabeled instances in the bags. In
order to achieve the goal, assumptions or rules are introduced to establish
relationship between labels of instances and the corresponding bag. Most MIL
models assume that all negative bags contain only negative instances, and that
positive bags contain at least one positive instance. However, there are also
other rules of the bag label definitions \cite{Srinidhi-etal-21}.

There are many MIL models which try to solve the problem under different
conditions and for different types of datasets
\cite{Andrews-etal-02,Chevaleyre-Zucker-01,Kraus-Ba-Frey-16,Sun-Han-etal-16,Wang-Yan-etal-18,Wang-Zucker-00}%
. Most above models use such methods as SVM, K nearest neighbors,
convolutional neural networks, decision trees. An interesting and efficient
class of the MIL models applies the attention mechanism
\cite{Pappas-PopescuBelis-17,Fuster-etal-21,Ilse-etal-18,Jiang-etal-21,Konstantinov-Utkin-22f,Rymarczyk-etal-21,Wang-Zhou-20}%
.

However, there are some disadvantages of the above approaches to MIL. First,
simple models based on such methods as SVM, decision trees, etc. do not use
neural network and cannot have several advantages of the network models, for
example, the end-to-end training and the attention mechanism. On the other
hand, the MIL models based on neural networks cannot be accurately trained on
small tabular datasets.

Therefore, the MIL model which simultaneously has properties of the random
forest (RF) and neural network can provide better results in comparison with
the available model. RF directly fit a MIL model because it is robust to noise
in target variables. At the same time, its structure is suboptimal, therefore,
it does not minimize the MIL loss in some cases. One of the ways to construct
decision trees, which can be retrained, is a concept of soft oblique trees
\cite{Heath-etal-93} whose trainable parameters can be updated and optimize by
using the gradient algorithms. Oblique trees and RFs composed from oblique
trees use linear and non-linear classifiers at each split in the decision
trees and allow us to combine more than one feature at a time. However, when
we deal with \emph{small tabular data}, soft oblique trees may overfit due to
a large number of parameters.

Therefore, we propose to represent soft oblique trees in the form of classic
decision trees and to convert the decision trees, which comprise RF, into the
trainable neural networks of a special way. The corresponding neural networks
implement approximately the same functions as the decision trees, but they can
be effectively trained jointly with the attention mechanism and can
simultaneously take into account data from all bags, i.e., they successfully
solve the MIL problem.

In sum, we propose an attention MIL model called Soft Tree Ensemble (STE-MIL).
On the one hand, STE-MIL is based on decision trees and successfully deals
with small tabular data. On the other hand, after converting trees into neural
networks and applying the attention mechanism to aggregate embedding of
instances and bags, STE-MIL is trained by using gradient descent algorithms in
an end-to-end manner.

Our contributions can be summarized as follows:

\begin{enumerate}
\item A new RF-based MIL model, which outperforms many MIL models when dealing
with small tabular data, is proposed.

\item A new type of soft decision trees similar to the soft oblique trees is
proposed. In contrast to the soft oblique trees, the proposed trees have a
smaller number of trainable parameters. Nevertheless, it can be trained in the
same way as the soft oblique trees. Outputs of each soft decision tree are
viewed as a set of vectors (embedding) which are formed from the class
probability distributions in a specific way.

\item An original algorithm for converting the decision trees into neural
networks of a specific form for efficient training parameters of the trees is proposed.

\item The attention is proposed to aggregate the instance and bag embeddings
with aim to minimize the corresponding loss function.

\item The whole MIL model, including soft decision trees, neural networks, the
attention mechanism and a classifier, is trained in an end-to-end manner.

\item Numerical experiments with well-known datasets Musk1, Musk2
\cite{Dietterich-etal-97}, Fox, Tiger, Elephant \cite{Andrews-etal-02}
illustrate STE-MIL. The above datasets have numerical features that are used
to perform tabular data. The corresponding code implementing the model is
publicly available at https://github.com/andruekonst/ste\_mil.
\end{enumerate}

The paper is organized as follows. Related work can be found in Section 2. A
brief introduction to MIL and the oblique binary soft trees is given in
Section 3. A specific representation of the decision tree function, which
allows us to convert the decision tree to a neural network, is proposed in
Section 4. A soft tree ensemble for solving the MIL problem is considered in
Section 5. An algorithm for converting the decision trees into neural networks
are studied in the same section. The attention mechanism applied to the
proposed MIL models is studied in Section 6. Numerical experiments are
provided in Section 7. Concluding remarks can be found in Section 8.

\section{Related work}

\textbf{MIL}. The MIL can be regarded as an important tool for dealing with
different types of data. In particular, tabular data of a specific structure
can be also classified by means of the MIL models. Therefore, if to consider
tabular data, then several available MIL models are based on the applying such
models as SVM, decision trees, AdaBoost, RFs
\cite{Andrews-etal-02,Chevaleyre-Zucker-01,Taser-etal-19,Wang-Yan-etal-18,Wang-Zucker-00}%
.

However, most MIL models are based on applying neural networks or
convolutional neural network, especially, when image datasets are classified
\cite{Doran-Ray-16,Feng-Zhou-17,Kraus-Ba-Frey-16,Liu-Zhou-etal-16,Sun-Han-etal-16,Wang-Yan-etal-18,Xu-16}%
.

In spite of various available MIL models, there are no models which could
combine the tabular data oriented models like RFs and neural networks,
including the attention mechanism, in order to use the gradient-based
algorithm for updating training parameters of RFs as well as neural networks
and to improve accuracy of the MIL predictions.

\textbf{MIL and attention.} Several MIL models using the attention mechanism
have been proposed in order to enhance the classification accuracy. Examples
of the models are SA-AbMILP (Self-Attention Attention-based MIL Pooling)
\cite{Rymarczyk-etal-21a}, ProtoMIL (Multiple Instance Learning with
Prototypical Parts) \cite{Rymarczyk-etal-21}, MHAttnSurv (Multi-Head Attention
for Survival Prediction) \cite{Jiang-etal-21}, AbDMIL \cite{Ilse-etal-18},
MILL (Multiple Instance Learning--based Landslide classification)
\cite{tang2021mill}, DSMIL (Dual-Stream Multiple Instance Learning)
\cite{Li-Li-Eliceiri-21}. The attention-based MIL models can be also found in
\cite{Fuster-etal-21,Pappas-PopescuBelis-17,Qi-etal-17,Schmidt-etal-23,Wang-Zhou-20}%
. The main peculiarity of the above mentioned models is that they use neural
networks and mainly deal with the image data, but not with small tabular data.

\textbf{Oblique trees and neural networks}. Many studies have demonstrated
that trees with oblique splits produce smaller trees with better accuracy
compared with axis parallel trees in many cases
\cite{Costa-Pedreira-22,Wickramarachchi-etal-16}. One of the important
advantages of oblique trees is that they can be trained by using optimization
algorithms, for example, the gradient descent algorithm. On the other hand,
some obstacles of training the oblique trees can be met. In particular, the
training procedure is computationally expensive. Moreover, the corresponding
model may be overfitted. Several approaches have been proposed to partially
solve the above problems. Wickramarachchi et al.
\cite{Wickramarachchi-etal-16} present a new decision tree algorithm, called
HHCART. In order to simplify oblique trees, Carreira-Perpinan and Tavallali
\cite{Carreira_Perpinan-Tavallali-18} propose an algorithm called sparse
oblique trees, which produces a new tree from the initial oblique tree having
the same or smaller structure, but new parameter values leading to a lower or
unchanged misclassification error. One-Stage Tree as a soft tree to build and
prune the decision tree jointly through a bi-level optimization problem is
presented in \cite{ZhuoerXu-etal-22}. Menze at al. \cite{Menze-etal-11} focus
on trees with task optimal recursive partitioning. Katuwal et al.
\cite{Katuwal-etal-20} propose a random forest of heterogeneous oblique
decision trees that employ several linear classifiers at each non-leaf node on
some top ranked partitions. An application of evolutionary algorithms to the
problem of oblique decision tree induction is considered in
\cite{Cantu_Paz-Kamath-03}. An algorithm improving learning of trees through
end-to-end training with backpropagation was presented in \cite{Hehn-etal-20}.

An interesting direction of using the oblique trees is to represent neural
networks as the trees or trees in the form of neural networks. Lee at al.
\cite{Lee-Jaakkola-19} show how neural models can be used to realize
piece-wise constant functions such as decision trees. Hazimeh et al.
\cite{Hazimeh-etal-20} propose to combine advantages of neural networks and
tree ensembles by designing a hybrid model by considering the so-called tree
ensemble layer for neural networks, which is an additive model of
differentiable decision trees. The layer can be inserted anywhere in a neural
network, and is trained along with the rest of the network using
gradient-based algorithms. Frosst and Hinton \cite{Frosst-Hinton-17} take the
knowledge acquired by a neural net and express the same knowledge in a model
that relies on hierarchical decisions instead, explaining a particular
decision would be much easier. A way of using a trained neural net to create a
type of soft decision tree that generalizes better than one learned directly
from the training data is also provided in \cite{Frosst-Hinton-17}.
Karthikeyan et al. \cite{karthikeyan2021learning} proposed a unified method
that enables accurate end-to-end gradient based tree training and can be
deployed in a variety of settings. Madaan et al. \cite{madaan2022treeformer}
presented dense gradient trees and an transformer based on the trees, which is
called Treeformer.

In contrast to the above work, we consider how to apply decision trees to the
MIL problem by converting the trees into neural networks of a special form and
by training them jointly with the attention mechanism in the end-to-end manner.

\section{Preliminary}

\subsection{Multiple Instance Learning}

First, we formulate the MIL classification problem. It differs from the
standard classification by the data structure. Namely, in the MIL problem,
bags have class labels, but instances, which compose each bags, are usually
unlabeled. This problem can be regarded as a kind of the weakly supervised
learning problem. Due to the availability of labels only for bags, the
following tasks can be stated in the framework of the MIL. The first task
concerns with annotation of instances from a bag. The second task aims to
classify new bags by having a training set of bags. The above tasks can be
solved by introducing special rules which establish the relationship between
the instance and bag class labels.

Let us formally state the MIL problem taking into account the rules connecting
different levels of the MIL data consideration. Suppose that each bag is
defined by a set of $m$ feature vectors $\mathbf{X}=\{\mathbf{x}%
_{1},...,\mathbf{x}_{n}\}$, where $\mathbf{x}_{i}\in\mathbb{R}^{m}$ is a
feature vector representing the $i$-th instance. Each instance $\mathbf{x}%
_{i}$ has a label $y_{i}\in\{0,1\}$ taking two values: $0$ (negative class)
and $1$ (positive class). We do not know labels $y_{i}$ during training as it
follows from the MIL problem statement. According to the first task, we have
to construct a function $g$ which maps each vector $\mathbf{x}_{i}$ into label
$y_{i}$.

There are various rules establishing the relationship between labels of bags
and instances. One of the most common rules can be rewritten as follows
\cite{Carbonneau-etal-18}:%
\begin{equation}
f(\mathbf{X})=\left\{
\begin{array}
[c]{cc}%
1, & \exists\mathbf{x}\in\mathbf{X}:g(\mathbf{x})=1,\\
0, & \text{otherwise,}%
\end{array}
\right.  \label{att_gbm_1}%
\end{equation}
where $f(\mathbf{X})$ is a bag classifier.

It follows from (\ref{att_gbm_1}) that at least one positive instance makes
the bag positive, and negative bags contain only negative instances. Function
$f(\mathbf{X})$ can be defined in another way taking into account a threshold
$\theta$ as%
\begin{equation}
f(\mathbf{X})=\left\{
\begin{array}
[c]{cc}%
1, & \theta\leq\sum_{\mathbf{x}\in\mathbf{X}}g(\mathbf{x}),\\
0, & \text{otherwise.}%
\end{array}
\right.
\end{equation}

We will use the rule defined in (\ref{att_gbm_1}).

The dataset can be represented as%

\begin{equation}
\mathcal{D}=\left\{  \left(  \{\mathbf{x}_{k}^{(i)}\}_{k=1}^{n_{i}}%
,y_{i}\right)  \right\}  _{i=1}^{N},
\end{equation}
where $\mathbf{x}_{k}^{(i)}$ is the $k$-th instance vector belonging to the
$i$-th bag; $n_{i}$ is the number of instances in the $i$-th bag; $N$ is the
number of labelled bags in the training set. \newline

Rule (\ref{att_gbm_1}) defining the function $f$ can be represented through
the MIL maximal pooling operator as follows:
\begin{equation}
f(\{\mathbf{x}_{k}^{(i)}\}_{k=1}^{n_{i}})=\max\left\{  g(\mathbf{x}_{k}%
^{(i)})\right\}  _{k=1}^{n_{i}}.
\end{equation}

Hence, the binary classification loss, which is minimized, can be written as
\begin{equation}
\mathcal{L}=\frac{1}{N}\sum_{i=1}^{N}l(y_{i},f(\{\mathbf{x}_{k}^{(i)}%
\}_{k=1}^{n_{i}}))=\frac{1}{N}\sum_{i=1}^{N}l(y_{i},\max\{g(\mathbf{x}%
_{k}^{(i)})\}_{k=1}^{n_{i}}). \label{eq:bag_loss}%
\end{equation}

\subsection{Oblique binary soft trees}

One of the important procedures to build oblique decision trees is
optimization of their parameters. There are various decision rules for
building trees. The so-called hard decision rules have been successfully
implemented in \cite{karthikeyan2021learning} and \cite{madaan2022treeformer}.
The rules are applied to the oblique decision trees which may be improper when
we deal with small tabular datasets, because large degrees of freedom in this
case would lead to overfitting.

According to \cite{karthikeyan2021learning}, an oblique binary tree of height
$h$ represents a piece-wise constant function $f(\mathbf{x};\mathbf{W}%
,\mathbf{b}):\mathbb{R}^{m}\rightarrow\mathbb{R}^{K}$ parameterized by weights
$\mathbf{w}_{I(d,l)}\in\mathbb{R}^{m}$, $b_{I(d,l)}\in\mathbb{R}$ at a node on
the path from the tree root to its leaf $l$ at depth $d$. Here $I(d,l)$ is the
index of a node on the path from the tree root to its leaf $l$ with depth $d$.
Function $f$ computes decision functions of the form $\mathbf{w}%
_{j}^{\mathrm{T}}\mathbf{x}-b_{j}>0$, that define whether $\mathbf{x}$ must
traverse the left or right child next. Here $\mathbf{W}$ is the parameter
matrix consisting of all parameter vectors $\mathbf{w}_{j}$; $\mathbf{b}$ is
the parameter vector consisting of parameters $b_{j}$. The tree output is
represented as $2^{h}$ vectors $\theta_{1},...,\theta_{2^{h}}$ such that
vector $\theta_{j}$ $\in\Delta^{K}$ at the $j$-th leaf is associated with
probabilities of $K$ classes, where $\mathbb{\Delta}^{K}$ is the unit simplex
of dimension $K$. One of the ways for learning parameters $\mathbf{w}_{ij}$
and $b_{ij}$ for all nodes is to minimize the expected loss $l$ of the form:
\begin{equation}
\min_{\mathbf{W},\mathbf{b}}\sum_{i=1}^{n}l(y_{i},f(\mathbf{x};\mathbf{W}%
,\mathbf{b})). \label{loss_func_oblique}%
\end{equation}

Karthikeyan et al. \cite{karthikeyan2021learning} propose the following
function $f(\mathbf{x};\mathbf{W},\mathbf{b})$:
\begin{equation}
f_{\theta}(\mathbf{x};\mathbf{W},\mathbf{b})=\sum_{l=1}^{2^{h}}q_{l}%
(\mathbf{x},\mathbf{W},\mathbf{b})\theta_{l}, \label{express_7}%
\end{equation}
where the tree path indicators $q_{l}(\mathbf{x},\mathbf{W},\mathbf{b})$ are
represented as the following indicator function:
\begin{equation}
q_{l}(\mathbf{x},\mathbf{W},\mathbf{b})=\mathbb{I}\left[  \bigwedge
\limits_{d=1}^{h}([\mathbf{w}_{I(d,l)}^{\mathrm{T}}\mathbf{x}\leq
b_{I(d,l)}]\oplus s(d,l))\right]  , \label{soft_tree_indicat_1}%
\end{equation}

Here $\mathbb{I}(\cdot)$ is the indicator function taking the value $1$ of its
argument is non-negative, otherwise it is $0$; $\oplus$ is the operator
\emph{XOR}; $s(d,l)$ determines whether the predicate of a node on the path to
the leaf $l$ at the depth $d$ should be evaluated to be true or false, i.e.,
$s(d,l)=1$ if the $l$-th leaf belongs to the left subtree of node $I(d,l)$,
otherwise $s(d,l)=0$. It is well known that the conjunction in
(\ref{soft_tree_indicat_1}) can be replaced with the product as follows:
\begin{equation}
q_{l}(\mathbf{x},\mathbf{W},\mathbf{b})=\prod_{d=1}^{h}\mathbb{I}\left[
[\mathbf{w}_{I(d,l)}^{\mathrm{T}}\mathbf{x}\leq b_{I(d,l)}]\oplus
s(d,l)\right]  .
\end{equation}

However, this representation significantly complicates the optimization of the
model by using the gradient descent algorithm due to the vanishing gradient
problem. Another representation of $q_{l}(\mathbf{x},\mathbf{W},\mathbf{b})$
is proposed in \cite{karthikeyan2021learning}. It is of the form:%

\begin{equation}
q_{l}(\mathbf{x},\mathbf{W},\mathbf{b})=\sigma\left(  \sum_{d=1}^{h}%
\sigma\left[  \lbrack\mathbf{w}_{I(d,l)}^{\mathrm{T}}\mathbf{x}\leq
b_{I(d,l)}]\oplus s(d,l)\right]  -h\right)  , \label{oblique2}%
\end{equation}
where the indicator functions are replaced with the so-called $\sigma$-hard
indicator approximations \cite{karthikeyan2021learning} which apply quantized
functions in the forward pass, but uses a smooth activation function in the
backward pass to propagate. This specific representation of the sigmoid
function is called the straight-through operator and proposed in
\cite{Bengio-etal-13}.

This representation allows us to effectively apply the gradient descent
algorithm to compute optimal parameters of the tree in accordance with the
loss function (\ref{loss_func_oblique}).

The soft tree concept proposed in \cite{karthikeyan2021learning} and
\cite{madaan2022treeformer} is an interesting approach to deal with small
tabular data. However, our experiments with soft trees have demonstrated that
it is difficult to train the oblique soft trees for some datasets. Therefore,
we propose to modify the standard decision trees to implement them in the form
of neural networks.

\section{A softmax representation of the decision tree function}

In order to overcome difficulties of training the oblique decision tree, we
propose another its representation which allows us to effectively update it.
Let us consider a complete binary decision tree $f_{\theta}$ of depth $h$:

\begin{itemize}
\item the tree has $(2^{h}-1)$ non-leaf nodes parametrized by $(\mathbf{w}%
_{j},b_{j})$, where

\begin{itemize}
\item $\mathbf{w}_{j}$ is an \emph{one-hot vector} having $1$ at the position
corresponding to the node feature;

\item $b_{j}$ is a threshold;
\end{itemize}

\item the tree has also $2^{h}$ leaves with values $\mathbf{v}_{l}$, where
$\mathbf{v}_{l}$ is an output vector corresponding to the $j$-th leaf.
\end{itemize}

In contrast to representation (\ref{oblique2}) of function $q_{l}$, we propose
to avoid the direct comparison with the height of a tree, because this
representation requires the indicator approximation to return integer values,
otherwise the $q_{l}$ will always be evaluated to be zero. If we would use
(\ref{oblique2}) instead of the softmax function, then (\ref{express_7}) will
provide the sum of the leaf vectors in place of selecting one of them. We use
the softmax function to guarantee a convex combination of the leaf vectors. We
replace the outer indicator with the \emph{softmax} function having the
trainable temperature parameter $\tau$:
\begin{equation}
q_{l}(\mathbf{x},\mathbf{W},\mathbf{b},\tau,\omega)=\mathrm{softmax}_{\tau
}\left(  \sum_{d=1}^{h}\sigma_{\omega}\left[  [-\mathbf{w}_{I(d,l)}%
^{\mathrm{T}}\mathbf{x}+b_{I(d,k)}]\cdot\hat{s}(d,k)\right]  \right)
_{k=1}^{2^{h}}, \label{softmax_tree_1}%
\end{equation}
where $\hat{s}(d,k)\in\{-1,1\}$ is the node sign; $\sigma$ is the sigmoid with
the trainable temperature or scaling parameter $\omega$.

The proposed representation could be interpreted as selecting the most
appropriate path among all candidate paths. Neural trees defined by using the
above representation can be optimized by means of the stochastic gradient
descent algorithm with the \emph{fixed node weights} $\mathbf{w}_{j}$, i.e.,
by updating only thresholds, the softmax temperature parameters $\tau$, the
sigmoid temperature parameters $\omega$ and leaf values.

\section{Soft Tree Ensemble for MIL}

One of the possible ways for solving the MIL classification problem, i.e., for
constructing the instance model $\tilde{g}$, is to assign a bag label to all
instances belonging to the bag. In this case, we get a new instance-level
dataset with the repeated instance labels, which is of the form:
\begin{equation}
\tilde{\mathcal{D}}=\{\left(  \mathbf{x}_{k}^{(i)},y_{i}\right)
~|~k=1,...,n_{i}\}_{i=1}^{N}. \label{eq:art_dataset}%
\end{equation}

According to \cite{leistner2010miforests}, RF can be regarded as a desirable
MIL classifier even if it is trained on artificially made instance-level
datasets like (\ref{eq:art_dataset}) because RF is inherently robust to noise
in the target variable. After training on dataset (\ref{eq:art_dataset}),
parameters of the built RF can be seen as a suboptimal solution to the
optimization problem defined by the bag-level loss (\ref{eq:bag_loss}). In the
extremely worst case, RF is totally overfitted, i.e. it just remembers the bag
label for each instance.

There are approaches that try to repeatedly infer the instance labels by using
a trained RF, and then retrain the RF on obtained instance labels. One of the
approaches is implemented in the so-called MIForests\emph{ }%
\cite{leistner2010miforests}. The main problem of the results is that the
methods rebuild decision trees instead of updating them, partially losing
useful tree structures obtained at different steps.

\subsection{Soft Tree Ensemble}

A key idea behind STE-MIL can be represented in the form of the following
schematic algorithm:

\begin{enumerate}
\item Let us assign incorrect labels to instances of a bag, for example, the
same as a label of the corresponding bag. The instance labels may be incorrect
because we do not know true labels and their determination is our task.
However, these labels are needed to build an initial RF. This is a kind of the
initialization procedure for the whole model which is trained in the
end-to-end manner.

\item The next step is to convert the initial RF to a neural network having a
specific architecture. To implement this step, non-leaf nodes of each tree in
RF are parametrized by trainable parameters $\mathbf{b}$, $\tau$, $\omega$,
and non-trainable parameters $\mathbf{W}$.

\item Parameters of the tree nodes $\mathbf{b}$, $\tau$, $\omega$ are updated
by using the stochastic gradient descend algorithm to minimize the bag loss
defined in (\ref{eq:bag_loss}). To implement the updating algorithm, we
propose to approximate the tree path indicators $q_{l}(\mathbf{x}%
,\mathbf{W},\mathbf{b},\tau,\omega)$ by using a specific softmax
representation (\ref{softmax_tree_1}). This is a key step of the algorithm
which allows us to update trees by updating neural networks and incorporate
the trees or RF in the whole scheme of modules, including the attention
mechanism and a classifier.
\end{enumerate}

Suppose that RF consisting of $T$ decision trees has been trained on repeated
instance labels (\ref{eq:art_dataset}). We convert its trees to a set of $T$
neural networks which implement functions $f^{(1)}(\mathbf{x}),\dots
,f^{(T)}(\mathbf{x})$ such that the $i$-th tree corresponds to the $i$-th
network implementing function $f^{(i)}(\mathbf{x})$. After converting trees to
neural networks, we can update their parameters to minimize the bag-level loss
(\ref{eq:bag_loss}). The ensemble prediction for a new instance $\mathbf{x}$
is defined as follows:
\begin{equation}
f(\mathbf{x})=\frac{1}{T}\sum_{i=1}^{T}f^{(i)}(\mathbf{x}).
\end{equation}
The bag prediction can be obtained by applying any aggregation function $G$:
\begin{equation}
f(\{\mathbf{x}_{k}^{(i)}\}_{i=1}^{N})=G(f(\mathbf{x}_{1}^{(i)}),\dots
,f(\mathbf{x}_{n_{i}}^{(i)})).
\end{equation}

The next questions is how to convert the decision trees into neural networks.

\subsection{Trees to Neural Networks}

Suppose that RF is trained on artificial dataset (\ref{eq:art_dataset}). Then
it can be converted to a neural network with a specific structure. A tree with
$M$ internal decision nodes and $L$ leaves is represented as a neural network
with three layers:

\begin{enumerate}
\item The first layer aims to approximate the node predicates. It is a fully
connected layer with $m$ inputs (dimensionality of $\mathbf{x}$) and $M$
outputs, i.e., there holds:
\begin{equation}
f^{(1)}(\mathbf{x})=\sigma\left(  \mathbf{Wx}+\mathbf{b}~|~\omega\right)  ,
\end{equation}
where $\mathbf{W}\in\mathbb{R}^{r\times m}$ is a matrix of non-trainable
parameters consisting of $r$ vectors $\mathbf{w}_{i}\in\mathbb{R}^{m}$; $r$ is
the total number of the tree nodes; $\mathbf{b}\in\mathbb{R}^{r}$ is the
trainable bias vector; $\omega$ is the trainable temperature parameter of the
sigmoid $\sigma$.

In sum, the first layer has only trainable parameters $\mathbf{b}$ and
$\omega$. The matrix $\mathbf{W}$ consists of the one-hot vectors having $1$
at positions corresponding to the node features.

\item The second layer aims to estimate the leaf indices. It is fully
connected layer with $M$ inputs and $L$ outputs having one trainable parameter
$\tau$:%
\begin{equation}
f^{(2)}(\xi)=\text{softmax}(\mathbf{R\xi}+\mathbf{s}~|~\tau),
\end{equation}
where $\mathbf{R}\in\mathbb{R}^{L\times M}$ is a non-trainable routing matrix
that encodes decision paths such that one path forms one row of $\mathbf{R}$;
$\mathbf{\xi}\in\mathbb{R}^{M}$ is the input vector; $\mathbf{s}\in
\mathbb{R}^{L}$ is the non-trainable bias vector; $\tau$ is the trainable
temperature parameter of the softmax operation.

Matrix $\mathbf{R}$ consists of values from the set: $\{-1,0,1\}$. If the path
to $i$-th leaf does not contain $j$-th node, then $R_{i,j}=0$. Otherwise, if
the path goes to the left branch, then $R_{i,j}=-1$, and $R_{i,j}=1$ if the
path goes to the right branch. Vector $\mathbf{s}=(s_{1},...,s_{2^{h}})$ is
needed to balance the decision paths. The sum of the sigmoid functions for the
path to the $k$-th leaf in (\ref{softmax_tree_1}) can be represented as:%
\begin{align}
&  \sum_{d=1}^{h}\sigma\left(  \lbrack-\mathbf{w}_{I(d,l)}^{\mathrm{T}%
}\mathbf{x}+b_{I(d,k)}]\cdot\hat{s}(d,k)\right) \nonumber\\
&  =\sum_{i=1}^{M}\left(  R_{k,i}\sigma\left(  -\mathbf{w}_{I(d,l)}%
^{\mathrm{T}}\mathbf{x}+b_{I(d,k)}\right)  +\mathbb{I}\left[  R_{k,i}%
=-1\right]  \right) \nonumber\\
&  =\sum_{i=1}^{M}\left(  R_{k,i}\sigma\left(  -\mathbf{w}_{I(d,l)}%
^{\mathrm{T}}\mathbf{x}+b_{I(d,k)}\right)  \right)  +s_{k},
\end{align}
because there holds $\sigma(-\omega)=1-\sigma(\omega)$.

\item The third layer aims to calculate the output values (embeddings). It is
trainable and fully connected. Each leaf generates the class probability
vector of size $C$. We take the probability $v_{1}(\mathbf{x})$ of class 1 and
repeat it $E-1$ times such that the whole embedding $\mathbf{v}(\mathbf{x}%
)=(v_{1}^{(1)}(\mathbf{x}),...,v_{1}^{(E)}(\mathbf{x}))$ has the length $E$.
The final output of the network (or the third layer) is of the form%
\begin{equation}
f(\mathbf{x})=\mathbf{V}f^{(2)}\left(  f^{(1)}(\mathbf{x})\right)  ,
\end{equation}
where $\mathbf{V}\in\mathbb{R}^{E\times L}$ is a trainable leaf value matrix
consisting of $L$ vectors $\mathbf{v}(\mathbf{x})$.
\end{enumerate}

\begin{figure}
[ptb]
\begin{center}
\includegraphics[
height=1.9112in,
width=2.2632in
]%
{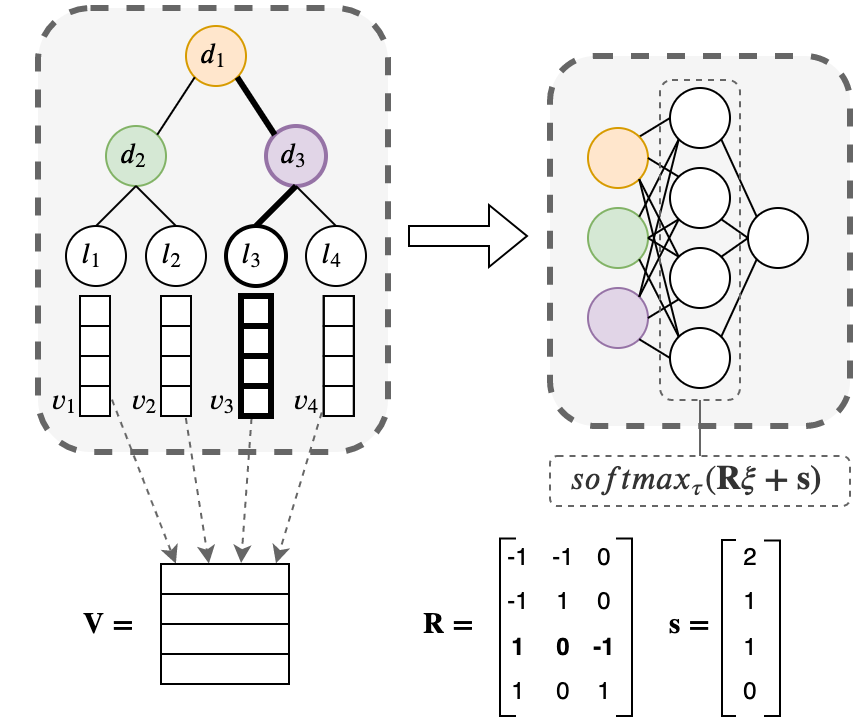}%
\caption{Forming the second layer through matrix $\mathbf{R}$}%
\label{fig:tree_to_nn}%
\end{center}
\end{figure}

An example of the transformation of a tree to a neural network is illustrated
in Fig. \ref{fig:tree_to_nn}. The full decision tree with three decision nodes
and four leaves is considered and depicted in Fig. \ref{fig:tree_to_nn}. The
first layer of the neural network computes all decisions at internal nodes of
the tree. Matrix $\mathbf{R}$ is constructed such that each its row represents
a path to the corresponding leaf of the tree. For example, values of the first
row are $(-1,-1,0)$ because the path to the leaf $l_{1}$ passes through the
nodes $d_{1}$ and $d_{2}$ to the left. Values of the third row are $(1,0,-1)$
because the path to the third leaf $l_{3}$ passes through the node $d_{1}$ to
the right, and it does not pass through the node $d_{2}$ and passes through
the node $d_{3}$ to the left. Elements of vector $\mathbf{s}$ are equal to
number of left turns, which is equivalent to the number of values $-1$ at the
corresponding row of $\mathbf{R}$.%

The class distribution provided by a tree is computed by counting the
percentage of different classes of instances at the leaf node where the
concerned instance falls into. Formally, the leaf value vectors are initially
estimated for the $l$-th leaf of the $j$-tree as follows:
\begin{equation}
\mathbf{v}_{j}^{(l)}(\mathbf{x})=\left(  \frac{\#\{(k,i)\in J(l)|y_{i}%
=1\}}{\#J(l)}\right)  _{t=1}^{E},
\end{equation}
where $J(l)$ is an index set of training points which fall into the $l$-th leaf.

We use constant matrix $\mathbf{W}$ that represents decision nodes, in order
to preserve the axis parallel decision planes. It is initialized as the
one-hot encoded representation of decision tree split features. Only the bias
$\mathbf{b}$ of the first layer of the neural network is trainable, and is
initialized with the negative values of the decision tree split thresholds.

Matrix $\mathbf{V}$ of the leaf values is initialized with repeated tree leaf
values, i.e., each its column contains the same values equal to the original
tree leaf value.

The algorithm for the routing matrix $\mathbf{R}$ construction is shown as
Algorithm \ref{alg:routing}.

\begin{algorithm}
\caption{Recursive $\bf{R}$ matrix construction}\label{alg:routing}
\begin{algorithmic}
\Procedure{FILL}{$\bf{R}$, $a$, $b$, $k$} \If{$k>M$} \Return
\EndIf
\State $d\leftarrow\lfloor\frac{b-a}{2}\rfloor$ \Comment{A half of the input row index span $[a, b]$}
\For{$i \gets a$ to $a + d$}
\State $R_{i,k}\leftarrow-1$ \Comment{Fill first $d$ rows with $-1$}
\State $R_{i,d+k}\leftarrow1$ \Comment{Fill second $d$ rows with $1$}
\EndFor
\State  \Call{FILL}{$\bf{R}$, $a$, $a + d$, $2 k$} \Comment{Recursively fill the first $d$ rows, left subtree}
\State \Call{FILL}{$\bf{R}$, $a + d$, $b$, $2 k + 1$} \Comment{Fill the second $d$ rows, right subtree}
\EndProcedure
\State $\mathbf{R}\leftarrow\mathbf{0}\in\mathbb{R}^{L\times M}$ \Comment{Initialize matrix with zeros}
\State \Call{FILL}{$\bf{R}$, $a=1$, $b=L$, $k=1$}
\end{algorithmic}
\end{algorithm}

\subsection{Peculiarities of the proposed soft trees}

\begin{itemize}
\item The sigmoid and softmax temperature parameters are trained starting from
value $0.1$ to avoid having to fit them as hyperparameters. Temperatures as
trainable parameters are not redundant because the first layer of the neural
network contains a fixed weight matrix $\mathbf{W}$, so $\mathbf{Wx}%
+\mathbf{b}$ cannot be equivalent to $\tau(\mathbf{Wx}+\mathbf{b})$. The same
take place with the softmax operation which contains a fixed number of terms
from $0$ to $1$.

\item In contrast to \cite{karthikeyan2021learning}, we do not use oblique
trees as they may lead to overfitting on tabular data. Trees with the
axis-parallel separating hyperplanes allow us to build accurate models for
tabular data where linear combinations of features often do not make sense.

\item Therefore, we also do not use overparametrization, which is a key
element for convergence of training the decision trees with quantized decision
rules (when the indicator is represented not by a sigmoid function, but by the
so-called straight-through operator \cite{Bengio-etal-13}).

\item We use softmax as an approximation of argmax instead of the
approximation of the sum of indicator functions. At the prediction stage, the
implementation of the algorithm proposed in \cite{karthikeyan2021learning},
which uses the sigmoid function, could predict the sum of the values at
several leaves at the same time.
\end{itemize}

Further, we can reduce the temperature $\omega$ so that the decision rules
become more stringent. Unfortunately, it is not working in practice because,
by a rather large depth ($h>3$), on the same path, inconsistent rules are
often learned, which give the \textquotedblleft correct\textquotedblright%
\ values by low temperatures and degenerate by small $\omega$. As a result,
the accuracy starts to decrease as $\omega$ decreases. If we do not decrease
$\omega$, then the trees may become not axis-parallel.

\section{Attention and the whole scheme of STE-MIL}

After training, output of each neural network corresponding to the $k$-th tree
is the embedding $\mathbf{v}_{j,k}^{(i)}$ of length $E$, where $i$ and $j$ are
indices of the corresponding bag and instance in the bag, respectively. This
implies that we get $T$ embeddings $\mathbf{v}_{j,1}^{(i)},...,\mathbf{v}%
_{j,T}^{(i)}$ for the $j$-th instance from the $i$-th bag, $j=1,...,n$,
$i=1,...,N$, under assumption of the identical number of trees in all RFs. It
should be noted that numbers of trees in RFs can be different. However, we
consider the same numbers for simplicity.

Embeddings $\mathbf{v}_{i,1},...,\mathbf{v}_{i,T}$ are aggregated by using,
for example, the averaging operation, resulting vectors $\mathbf{e}_{j}^{(i)}%
$, $j=1,...,n$, corresponding to the $i$-th bag. Then aggregated embeddings
$\mathbf{e}_{1}^{(i)},...,\mathbf{e}_{n}^{(i)}$ are attended in order to get a
final representation of the $i$-th bag in the form of vector $\mathbf{a}_{i}$,
which is classified. This motivates us to replace the class probability
distributions at the tree leaves with the embeddings $\mathbf{v}$ defined
above. We can define several ways for constructing embeddings from the class
probability distributions. However, we select a simple procedure which has
demonstrated its efficiency from the accuracy and computational points of view.

\begin{figure}
[ptb]
\begin{center}
\includegraphics[
height=2.092in,
width=5.1154in
]%
{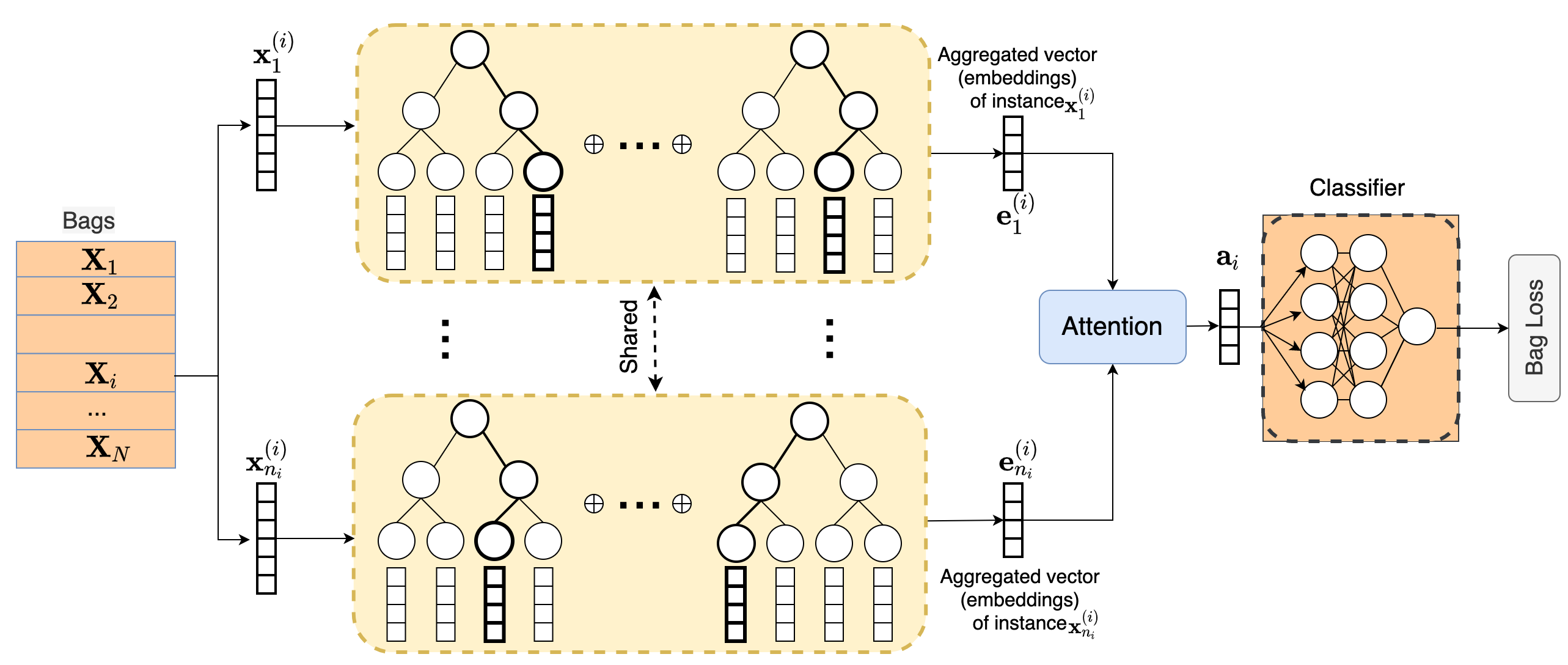}%
\caption{A scheme of the ensembled STE-MIL}%
\label{f:ste_scheme}%
\end{center}
\end{figure}

Hence, the second idea behind STE-MIL is to aggregate the embeddings\ over all
bags by using the attention mechanism and to calculate the prediction logits
by linear projecting the aggregated embedding to the one-dimensional space.
This idea is also motivated by the Attention-MIL approach proposed in
\cite{Ilse-etal-18} and by the Multi-attention multiple instance learning
model proposed in \cite{Konstantinov-Utkin-22f}, which may help to train a
better bag-level classifier. A scheme of the whole STE-MIL model is shown in
Fig. \ref{f:ste_scheme}. It can be seen from Fig. \ref{f:ste_scheme} that each
instance ($\mathbf{x}_{j}^{(i)}$) from the $i$-th bag learns the corresponding
RF such that embeddings are combined to the aggregated vector $\mathbf{e}%
_{j}^{(i)}$. Vectors $\mathbf{e}_{j}^{(i)}$, $j=1,...,n_{i}$, can be regarded
as keys in terms of the attention and are attended and produce vector
$\mathbf{a}_{i}$ which is the input of the classifier. The whole system is
trained on all instances from all bags.

The attention module produces a new aggregate embedding $\mathbf{a}_{k}$
corresponding to the $k$-th bag, which is computed as follows:%
\begin{equation}
\mathbf{a}_{k}=\sum_{i=1}^{n}\beta_{i}^{(k)}\mathbf{e}_{i}^{(k)},\ k=1,...,N,
\end{equation}
where
\begin{equation}
\beta_{i}^{(k)}=\mathrm{softmax}\left(  \mathbf{q}^{\mathrm{T}}\mathbf{k}%
_{i}\right)  ,
\end{equation}%
\begin{equation}
\mathbf{q=V}_{q}\mathbf{g},\ \mathbf{k}_{i}\mathbf{=V}_{k}\mathbf{e}_{i}%
^{(k)}.
\end{equation}

Here $\mathbf{V}_{k}$ and $\mathbf{V}_{q}$ are the trainable weight matrices
for $\mathbf{e}_{i}^{(k)}$ (keys) and the template vector $\mathbf{g}$
(query), respectively.%

\section{Numerical experiments}

In order to compare the proposed model with other MIL classification models,
we train the corresponding models on datasets Musk1, Musk2 (drug activity)
\cite{Dietterich-etal-97}, Fox, Tiger, Elephant \cite{Andrews-etal-02}. Table
\ref{t:class_datasets} shows the number of bags $N$, the number of instances
$n$ in every bag and the number of features $m$ in instances for the
corresponding datasets. The Musk1 dataset contains 92 bags consisting of 476
instances with 166 features. The average bag size is $5.17$. The Musk2 dataset
contains 102 bags consisting of 6598 instances with 166 features. The average
bag size is $64.69$. Each dataset (Fox, Tiger and Elephant) contains exactly
200 bags consisting of instances with 230 features. Numbers of instances in
datasets Fox, Tiger and Elephant are 1302, 1220 and 1391, respectively. The
average bag sizes of the datasets are $6.60$, $6.96$ and $6.10$, respectively.

The accuracy measures for these datasets are also obtained by means of the
well-known MIL classification models, including mi-SVM \cite{Andrews-etal-02},
MI-SVM \cite{Andrews-etal-02}, MI-Kernel \cite{Gartner-etal-02}, EM-DD
\cite{Zhang-Goldman-02}, mi-Graph \cite{Zhou-Sun-Li-09}, miVLAD
\cite{Wei-Wu-Zhou-17}, miFV \cite{Wei-Wu-Zhou-17}, mi-Net
\cite{Wang-Yan-etal-18}, MI-Net \cite{Wang-Yan-etal-18}, MI-Net with DS
\cite{Wang-Yan-etal-18}, MI-Net with RC \cite{Wang-Yan-etal-18}, Attention and
Gated-Attention \cite{Ilse-etal-18}.%

\begin{table}[tbp] \centering
\caption{A brief introduction about datasets for classification}%
\begin{tabular}
[c]{cccc}\hline
Data set & $N$ & $n$ & $m$\\\hline
Elephant & $200$ & $1391$ & $230$\\
Fox & $200$ & $1302$ & $230$\\
Tiger & $200$ & $1220$ & $230$\\
Musk1 & $92$ & $476$ & $166$\\
Musk2 & $102$ & $6598$ & $166$\\\hline
\end{tabular}
\label{t:class_datasets}%
\end{table}%

We investigate Extremely Randomized Trees (ERT) for initialization because
they provide better results. At each node, the ERT algorithm chooses a split
point randomly for each feature and then selects the best split among these
\cite{Geurts-etal-06}.

We also use in experiments: sigmoid function with the trainable temperature
parameter $\omega$, which is initialized with $10$, as indicator
approximation; softmax operation with the trainable temperature parameter
$\tau$, which is is also initialized with $10$; the number $T$ of decision
trees is $20$; the largest depth $h$ of trees is $5$; the dimension $E$ of
each embedding vector is $4$; the number of epochs is $2000$; the batch size
is $20$ and the learning rate is $0.01$.

Accuracy measures (the mean and standard deviation) are computed by using
5-fold cross-validation. The best results in tables are shown in bold.
Numerical results for datasets Elephant, Fox and Tiger are shown in Table
\ref{t:strmil2}. It can be seen from Table \ref{t:strmil2} that STE-MIL
provides outperforming results for all datasets. Numerical results for
datasets Musk1 and Musk2 are shown in Table \ref{t:strmil3}. One can see from
Table \ref{t:strmil3} that the proposed model outperforms all other models for
the dataset Musk1. However, STE-MIL provides the worse result for the dataset
Musk2. One of the reasons of this result is that bags in Musk2 consist of many
instances. This implies that the advantage of STE-MIL to deal with small
datasets cannot be shown on this dataset.%

\begin{table}[tbp] \centering
\caption{Accuracy measures (the mean and standard deviation) for comparison of the well-known MIL classification models, the RF and the STE-MIL by using datasets Elephant, For and Tiger}%
\begin{tabular}
[c]{llll}\hline
& Elephant & Fox & Tiger\\\hline
mi-SVM \cite{Andrews-etal-02} & \multicolumn{1}{r}{0.822$\pm$N/A} &
\multicolumn{1}{r}{0.582$\pm$N/A} & \multicolumn{1}{r}{0.784$\pm$N/A}\\
MI-SVM \cite{Andrews-etal-02} & \multicolumn{1}{r}{0.843$\pm$N/A} &
\multicolumn{1}{r}{0.578$\pm$N/A} & \multicolumn{1}{r}{0.840$\pm$N/A}\\
MI-Kernel \cite{Gartner-etal-02} & \multicolumn{1}{r}{0.843$\pm$N/A} &
\multicolumn{1}{r}{0.603$\pm$N/A} & \multicolumn{1}{r}{0.842$\pm$N/A}\\
EM-DD \cite{Zhang-Goldman-02} & \multicolumn{1}{r}{0.771$\pm$0.097} &
\multicolumn{1}{r}{0.609$\pm$0.101} & \multicolumn{1}{r}{0.730$\pm$0.096}\\
mi-Graph \cite{Zhou-Sun-Li-09} & \multicolumn{1}{r}{0.869$\pm$0.078} &
\multicolumn{1}{r}{0.620$\pm$0.098} & \multicolumn{1}{r}{0.860$\pm$0.083}\\
miVLAD \cite{Wei-Wu-Zhou-17} & \multicolumn{1}{r}{0.850$\pm$0.080} &
\multicolumn{1}{r}{0.620$\pm$0.098} & \multicolumn{1}{r}{0.811$\pm$0.087}\\
miFV \cite{Wei-Wu-Zhou-17} & \multicolumn{1}{r}{0.852$\pm$0.081} &
\multicolumn{1}{r}{0.621$\pm$0.109} & \multicolumn{1}{r}{0.813$\pm$0.083}\\
mi-Net \cite{Wang-Yan-etal-18} & \multicolumn{1}{r}{0.858$\pm$0.083} &
\multicolumn{1}{r}{0.613$\pm$0.078} & \multicolumn{1}{r}{0.824$\pm$0.076}\\
MI-Net \cite{Wang-Yan-etal-18} & \multicolumn{1}{r}{0.862$\pm$0.077} &
\multicolumn{1}{r}{0.622$\pm$0.084} & \multicolumn{1}{r}{0.830$\pm$0.072}\\
MI-Net with DS \cite{Wang-Yan-etal-18} & \multicolumn{1}{r}{0.872$\pm$0.072} &
\multicolumn{1}{r}{0.630$\pm$0.080} & \multicolumn{1}{r}{0.845$\pm$0.087}\\
MI-Net with RC \cite{Wang-Yan-etal-18} & \multicolumn{1}{r}{0.857$\pm$0.089} &
\multicolumn{1}{r}{0.619$\pm$0.104} & \multicolumn{1}{r}{0.836$\pm$0.083}\\
Attention \cite{Ilse-etal-18} & \multicolumn{1}{r}{0.868$\pm$0.022} &
\multicolumn{1}{r}{0.615$\pm$0.043} & \multicolumn{1}{r}{0.839$\pm$0.022}\\
Gated-Attention \cite{Ilse-etal-18} & \multicolumn{1}{r}{0.857$\pm$0.027} &
\multicolumn{1}{r}{0.603$\pm$0.029} & \multicolumn{1}{r}{0.845$\pm$0.018}\\
\textbf{STE-MIL} & \multicolumn{1}{r}{\textbf{0.885}$\pm$0.038} &
\multicolumn{1}{r}{\textbf{0.730}$\pm$0.080} &
\multicolumn{1}{r}{\textbf{0.875}$\pm$0.039}\\\hline
\end{tabular}
\label{t:strmil2}%
\end{table}%
%

\begin{table}[tbp] \centering
\caption{Accuracy measures (the mean and standard deviation) for comparison of the well-known MIL classification models, the RF and the STE-MIL by using datasets Musk1 and Musk2 }%
\begin{tabular}
[c]{lll}\hline
& Musk1 & Musk2\\\hline
mi-SVM \cite{Andrews-etal-02} & \multicolumn{1}{r}{0.874$\pm$N/A} &
\multicolumn{1}{r}{0.836$\pm$N/A}\\
MI-SVM \cite{Andrews-etal-02} & \multicolumn{1}{r}{0.779$\pm$N/A} &
\multicolumn{1}{r}{0.843$\pm$N/A}\\
MI-Kernel \cite{Gartner-etal-02} & \multicolumn{1}{r}{0.880$\pm$N/A} &
\multicolumn{1}{r}{0.893$\pm$N/A}\\
EM-DD \cite{Zhang-Goldman-02} & \multicolumn{1}{r}{0.849$\pm$0.098} &
\multicolumn{1}{r}{0.869$\pm$0.108}\\
mi-Graph \cite{Zhou-Sun-Li-09} & \multicolumn{1}{r}{0.889$\pm$0.073} &
\multicolumn{1}{r}{\textbf{0.903}$\pm$0.086}\\
miVLAD \cite{Wei-Wu-Zhou-17} & \multicolumn{1}{r}{0.871$\pm$0.098} &
\multicolumn{1}{r}{0.872$\pm$0.095}\\
miFV \cite{Wei-Wu-Zhou-17} & \multicolumn{1}{r}{0.909$\pm$0.089} &
\multicolumn{1}{r}{0.884$\pm$0.094}\\
mi-Net \cite{Wang-Yan-etal-18} & \multicolumn{1}{r}{0.889$\pm$0.088} &
\multicolumn{1}{r}{0.858$\pm$0.110}\\
MI-Net \cite{Wang-Yan-etal-18} & \multicolumn{1}{r}{0.887$\pm$0.091} &
\multicolumn{1}{r}{0.859$\pm$0.102}\\
MI-Net with DS \cite{Wang-Yan-etal-18} & \multicolumn{1}{r}{0.894$\pm$0.093} &
\multicolumn{1}{r}{0.874$\pm$0.097}\\
MI-Net with RC \cite{Wang-Yan-etal-18} & \multicolumn{1}{r}{0.898$\pm$0.097} &
\multicolumn{1}{r}{0.873$\pm$0.098}\\
Attention \cite{Ilse-etal-18} & \multicolumn{1}{r}{0.892$\pm$0.040} &
\multicolumn{1}{r}{0.858$\pm$0.048}\\
Gated-Attention \cite{Ilse-etal-18} & \multicolumn{1}{r}{0.900$\pm$0.050} &
\multicolumn{1}{r}{0.863$\pm$0.042}\\
\textbf{STE-MIL} & \multicolumn{1}{r}{\textbf{0.918}$\pm$0.077} &
\multicolumn{1}{r}{0.854$\pm$0.061}\\\hline
\end{tabular}
\label{t:strmil3}%
\end{table}%

\section{Conclusion}

A RF-based model for solving the MIL classification problem for small tabular
data has been proposed. It is based on training decision trees by means of
their converting to neural network of a specific form. Moreover, it uses the
attention mechanism to aggregate the bag information and to enhance the
classification accuracy. The attention mechanism can also be used to explain
why a tested bag is assigned by a certain label because the attention shows
weights of instances of the tested bag and selects the most influential instances.

Numerical experiments with the well-known datasets, which are used by many
authors for evaluating the MIL models, have demonstrated that STE-MIL
outperforms many models, including mi-SVM, MI-SVM, MI-Kernel, EM-DD, mi-Graph,
miVLAD, miFV, mi-Net, MI-Net, MI-Net with DS, MI-Net with RC, the Attention
and Gated-Attention models, for most datasets analyzed.

The main advantage of STE-MIL is that it opens a door for constructing various
models which use trainable decision trees as neural networks. In contrast to
models using oblique decision trees, the proposed trainable tress have the
significantly small number of training parameters preventing overfitting of
the training process. Therefore, these models could be effective when small
tabular datasets are considered.

Ideas of STE-MIL can be used in other known MIL models. For example, it is
interesting to incorporate the neighboring patches or instances of each
analyzed patch into the STE-MIL scheme as it has been made in
\cite{Konstantinov-Utkin-22f}. The incorporation of neighbors can
significantly improve STE-MIL. This can be viewed as a direction for further research.

It should be noted that RF as an ensemble of decision trees has been used in
STE-MIL. However, the gradient boosting machine
\cite{Friedman-2001,Friedman-2002} is also an efficient model which uses
decision trees as weak learners. Therefore, another idea behind new models is
to apply the gradient boosting machine. This is another direction for further research.

\bibliographystyle{unsrt}
\bibliography{Attention,MIL,Deep_Forest,MYBIB,Boosting}

\end{document}